\documentclass[a4paper,fleqn]{cas-dc}
\usepackage[numbers]{natbib}
\makeatletter
\let\origsection\section
\renewcommand\section{\nostarsection}
\renewcommand\section{\@ifstar{\starsection}{\nostarsection}}
\newcommand\nostarsection[1]{\vspace{-0.5em}\origsection{#1}}
\newcommand\starsection[1]{\vspace{-0.5em}\origsection*{#1}}
\makeatother

\usepackage{float}
\newcommand{\eref}[1]{(\ref{#1})}
\newcommand{\fref}[1]{Figure~\ref{#1}}
\newcommand{\E}[1]{{\mathrm{E}}\!\left[#1\right]}
\newcommand{\EE}[2]{{\mathrm{E}_{#1}}\!\left[#2\right]}
\newcommand{\df}{\mathrm{d}}
\newcommand{\bup}{\pmb \upsilon}
\newcommand{\up}{\upsilon}
\newcommand{\eee}{\mathrm{e}}

\begin{document}

\let\WriteBookmarks\relax
\def\floatpagepagefraction{1}
\def\textpagefraction{.001}
\shorttitle{K. Atitey et al./Variational Bayesian inference of hidden
  processes}
\shortauthors{K. Atitey et~al.}

\title[mode = title]{Variational Bayesian inference of hidden stochastic
  processes\\ with unknown parameters}

\author[1]{Komlan Atitey} \cormark[1]\ead{atiteydavidkomlan@yahoo.fr} 
\author[1]{Pavel Loskot} \ead{p.loskot@swansea.ac.uk}
\address[1]{Swansea University, College of Engineering, Fabian Way, Skewen,
  Swansea SA1 8EN, United Kingdom}

\author[2] {Lyudmila Mihaylova}\ead{L.S.Mihaylova@sheffield.ac.uk}
\address[2]{University of Sheffield, Dept. of Automatic Control and
  Systems Engineering, Mappin Street, Sheffield S1 3JD, United Kingdom}

\cortext[cor1]{Corresponding author}

\begin{abstract}
  Estimating hidden processes from non-linear noisy observations is
  particularly difficult when the parameters of these processes are not known.
  This paper adopts a machine learning approach to devise variational Bayesian
  inference for such scenarios. In particular, a random process generated by
  the autoregressive moving average (ARMA) linear model is inferred from
  non-linearity noisy observations. The posterior distributions of hidden
  states are approximated by a set of weighted particles generated by the
  sequential Monte Carlo (SMC) algorithm involving sampling with importance
  sampling-resampling (SISR). Numerical efficiency and estimation accuracy of
  the proposed inference method are evaluated by computer simulations.
  Furthermore, the proposed inference method is demonstrated on a practical
  problem of estimating the missing values in the gene expression time series
  assuming vector autoregressive (VAR) data model.
\end{abstract}


\maketitle

\section{Introduction}

Filtering stochastic processes and time series is the important problem in
scientific data processing. This task is particularly difficult when the
underlying data model is non-linear, and its parameters are unknown. Moreover,
the posterior distribution of hidden samples may need to be updated
sequentially as new observations arrive. A common strategy to address the
computational complexity of Bayesian inference is to assume approximations of
the posterior distribution. For instance, the approximation utilizing the
Markov Chain Monte Carlo (MCMC) sampling is adopted in
\cite{durlauf2016macroeconometrics, green2015bayesian}. Variational Bayesian
inference for non-linear models is investigated in \cite{blei2017variational,
  raiko2007building}. In \cite{helske2017computational}, Bayesian inference is
implemented assuming an asymptotically exact MCMC pseudo-marginal particle
filter, and also employing the extended and unscented Kalman filters. However,
these inference methods are still very computationally demanding, and properly
setting the parameters of these algorithms is not an easy task.

The time series data can be modeled using a non-linear discrete time ARMA model
\cite{kamarianakis2003spatial}. Inspired by the problems in statistical
physics, the latent states of non-linear ARMA model with unknown parameters
were estimated by the importance sampling sequential Monte Carlo (IS-SMC)
method and the density assisted sequential Monte Carlo (DA-SMC) method in
\cite{urteaga2017sequential}. The former method selects the joint proposal
density before applying the IS to generate samples of latent states and of
unknown parameters. The latter method approximates the posterior of unknown
parameters by the Gaussian distribution. However, the accuracy achieved by
these methods was not satisfactory. Variational Bayesian inference of latent
states is studied in \cite{acerbi2018variational, daunizeau2009variational},
and modified in \cite{blei2017variational, raiko2007building} by assuming the
joint density of latent states and of an auxiliary random variable. The
auxiliary random variable transforms the linear ARMA model into a stochastic
volatility model driven by the fractional Gaussian process.

In this paper, our objective is to investigate Bayesian inference of discrete
time random processes with unknown parameters under non-linear and noisy
observations. We specifically consider random processes generated by the ARMA
models as they are frequently studied in the literature
\cite{beal2003variational, geiger2015optimal}. However, we allow the underlying
ARMA model to be driven by a fractional Gaussian process. We adopt variational
Bayesian inference, and to make the computational complexity tractable, it is
implemented as the SMC-SISR estimation. This estimator numerically approximates
the posterior density by a set of weighted samples referred to as particles.
The estimator accuracy is evaluated by computer simulations. In addition, the
developed estimator is used to infer the missing values in the gene expression
time series data modeled as a vector autoregressive (VAR) process
\cite{roman2018mutant, tam2013synthetic}.

The rest of this paper is organized as follows. Section 2 describes a
non-linear ARMA random process. Variational Bayesian inference and the SMC-SISR
estimator are developed in Section 3. Numerical examples are presented in
Section 4 followed by a practical problem of inferring missing values in the
time series data. The results are discussed in Section 6. Section 7 concludes
the paper.

\section{Non-linear observation model of a hidden random process}

A canonical time-invariant ARMA$(m, n)$ process of orders $m$ and $n$ is
defined by its $m$ autoregressive (AR) coefficients $\{\phi_{1:m}\}$, and $n$
moving average (MA) coefficients $\{\varphi_{1:n}\}$. These parameters define
the autocorrelation of the output random process $x_t$ generated from the input
innovations $u_t$ at discrete time instances $t$. A non-linear ARMA model is
obtained by introducing a non-linearity $g$ with the observation noise $\up_t$,
i.e., the noisy observations $z_t$,
\begin{eqnarray}\label{eq:10}
  x_t &=& \sum_{i=1}^{m} {\phi_i}{x_{t-i}} +\sum_{j=1}^{n}
          {\varphi_j}{u_{t-j}}+u_t   \\
  z_t & =& g(x_t,\up_t). \nonumber
\end{eqnarray}
Model \eref{eq:10} is also a state-space description of dynamic systems
\cite{durlauf2016macroeconometrics, liu2016online}. The innovations $u_t$ are
assumed to be a zero-mean stationary Gaussian process. The non-linearity $g$ is
constrained by the requirement that the likelihood $p(z_t|x_t)$ of $x_t$ is
computable up to a proportionality constant. In matrix notation, model
\eref{eq:10} until current time $t$ can be rewritten as,
\begin{equation*}
  \pmb \Phi_t \pmb x_t=\pmb \Psi_t \pmb u_t
\end{equation*}
where the vectors $\pmb x_t = \{ x_{1:t} \} $ and $ \pmb u_t = \{ u_{1:t} \} $,
and the transition matrices,
\begin{equation*}    
  \setlength{\arraycolsep}{2pt}
  \renewcommand{\arraystretch}{0.8}
  \pmb \Phi_t = 
  \begin{pmatrix}
    1 & -\phi_1 & \cdots & - \phi_m & \cdots & 0 \\
    & 1 & -\phi_1 & \cdots & -\phi_m & \vdots \\
    & & \ddots & & \ddots & \\
    & & & 1 & -\phi_1 & -\phi_2 \\
    \vdots & & & & 1 & -\phi_1 \\
    0 & \cdots & & & & 1 
  \end{pmatrix}
\end{equation*}
\begin{equation*}
  \pmb \Psi_t = 
  \begin{pmatrix}
    1 & \varphi_1 & \cdots & \varphi_n & \cdots & 0 \\
    & 1 & \varphi_1 & \cdots & \varphi_n & \vdots \\
    & & \ddots & & \ddots & \\
    & & & 1 & \varphi_1 & \varphi_2 \\
    \vdots & & & & 1 & \varphi_1 \\
    0 & \cdots & & & & 1 
  \end{pmatrix}.
\end{equation*}
Consequently, the state vector $\pmb x_t$ can be expressed as the linear
transformation of innovations \cite{nogales2006electricity}, i.e.,
\begin{equation}\label{eq:25}
  \pmb x_t= \pmb \Phi_t^{-1} \pmb \Psi_t \pmb u_t = \pmb \Theta_t \pmb u_t.
\end{equation}
The transfer matrix $\pmb \Theta_t$ determines the observability of innovations
from the states $\pmb x_t$ as well as controlability of the ARMA model
\cite{mills1991time, nogales2006electricity, tsay2005analysis}.

Even though the Gaussian innovation process $u_t$ is normally assumed to be
white, i.e., its samples are uncorrelated, in this paper, we assume the
autocorrelation,
\begin{eqnarray}
  \gamma_{\pmb u}(\tau) &=& \E{u_{t+\tau} u_t^\ast} =
   \sigma_u^2 \,\rho_{\pmb u}(\tau) \\ &=& \frac{\sigma_u^2}{2}
       \Big[|\tau + 1|^{2H} - 2|\tau|^{2H} + |\tau - 1|^{2H}\Big] \nonumber
\end{eqnarray}
of so called Gaussian fractional process where $0< H <1$ is the Hurst exponent,
and the variance $\sigma_u^2$ is set to obtain the normalization,
$\rho_{\pmb u}(0)=1$. For integer values $\tau$, the covariance matrix of the
zero-mean Gaussian vector $\pmb u_t$ can be expressed as,
\begin{equation}
  \pmb C_{\pmb u_t} = \sigma_u^2 \,\pmb R_{\pmb u_t}
\end{equation}
where
\begin{equation*}
    \pmb R_{\pmb u_t} = \begin{pmatrix}
    \rho_u(1) & \rho_u(2) & \cdots & \rho_u(t-1) \\
    \rho_u(2) & \rho_u(1) & \cdots & \rho_u(t-2) \\
    \vdots & \vdots & \ddots &  \vdots \\
    \rho_u(t-1) & \rho_u(t-2) & \cdots & \rho_u(1) \\
    \rho_u(t) & \rho_u(t-1) & \cdots & \rho_u(2) \\
  \end{pmatrix}.
\end{equation*}
The covariance matrix for the zero-mean Gaussian distributed states $x_t$ is
given as,
\begin{equation*}
  \pmb C_{\pmb x_t} = \pmb \Psi_t C_{\pmb u_t} \pmb \Psi_t^T =
  \sigma_u^2\, \pmb \Psi_t R_{\pmb u_t} \pmb \Psi_t^T
\end{equation*}
where $(\cdot)^T$ denotes the matrix transpose. 

\subsection{Posterior distribution of states}

Assume for now that the matrices $\pmb \Phi_t$ and $\pmb \Psi_t$ are known.
Given non-linearity $g$, model \eref{eq:10} can be statistically described by
the transition probabilities $p(x_t|x_{t-1})$ and the likelihood $p(z_t|x_t)$,
and we assume that given states $x_t$, the observations $z_t$ are conditionally
independent. The joint posterior of states $\pmb x_t$ is given by the Bayes
theorem, i.e.,
\begin{equation}\label{eq:20}
  p(\pmb x_t|\pmb z_t) = \frac{p(\pmb z_t|\pmb x_t)p(\pmb x_t)}{ \int_{\pmb
      \Omega_x^t} p(\pmb z_t| \pmb x_t) p(\pmb x_t)\, \df \pmb x_t} 
\end{equation}
where $\pmb \Omega_x^t$ is the support of vector $\pmb x_t$. For long time
series data, the posterior \eref{eq:20} must be computed recursively and
numerically, for example, using Monte Carlo sampling methods. In order to
obtain such a method for recursively calculating the posterior of state $x_t$
from the previous states $\pmb x_{t-1}$ and the observations $\pmb z_t$, we
need the conditional probability predicting the current state $x_t$ from the
previous states $\pmb x_{t-1}$. This probability is computed by solving the
Chapman-Kolmogorov equation,
\begin{equation}
  p(x_t|\pmb z_{t-1}) = \int_{\pmb \Omega_x^{t-1}} p(x_t|\pmb
  x_{t-1}) p(\pmb x_{t-1}|\pmb z_{t-1})\, \df \pmb x_{t-1}.
\end{equation}

\subsection{Posterior distribution of states with unknown parameters}

Recall that our aim is to estimate the random states $x_t$ from observations
$z_t$. We now consider a common scenario that the parameters of ARMA model are
not known. Although the unknown parameters can be estimated before calculating
the posteriors of states $x_t$, these parameters may be time varying, and
difficult to estimate even for ARMA models with small orders. In order to avoid
challenges in explicitly estimating the ARMA model parameters, we need to
rewrite the expressions for the posterior distribution of states $x_t$
presented in the previous subsection.

\begin{figure}
  \centering
  \includegraphics[width=1\linewidth]{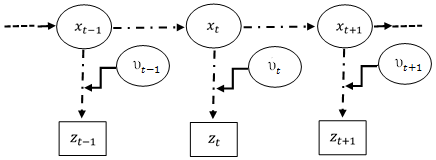}
  \caption{Causality of observations and of the latent variables.}
  \label{fig1}
\end{figure}

In particular, assume that the random variables $\up_t$ are independent and
identically distributed (IID). Since in many scenarios, the states $x_t$ are
non-negative integers such as the object counts, it is useful to constrain the
noises, so that $\upsilon_t\geq 0$. Among different probability distributions
with positive support, gamma distribution appears to be the most commonly
occurring, so we assume that the IID noise samples are gamma distributed
\cite{elster2015guide, leon2017probability}, i.e.,
\begin{equation}
  \up_t \sim G_\Upsilon(\up_t;\alpha,\beta)= \frac{\up_t^{\beta-1}}{\up_t^\beta
    \,\Gamma(\beta)} \eee^{-\up_t/\alpha}
\end{equation}
where $\Gamma$ denotes the gamma function, and $\alpha,\beta>0$ are the
parameters of the gamma distribution $G$. Since the noise samples are IID, we
can write,
\begin{equation}
  \bup_t \sim \prod_{l=1}^t G_\Upsilon(\up_l; \alpha,\beta).
\end{equation}
Thus, non-linear observations $z_t$ in \eref{eq:10} represent bivariate
transformation of the Gaussian and gamma distributed hidden random variables as
indicated in \fref{fig1}.

The joint probability density $p(\up_t,x_t|\pmb z_{t-1})$ can be expressed
recursively as \cite{atitey2015novel, vo2006gaussian},
\begin{equation}
  \begin{split}
    p(\up_t,x_t|\pmb z_{t-1}) =
    &\int_{\pmb \Omega_{\up,x}^{t-1}} p(\up_t,x_t|\pmb \up_{t-1},\pmb x_{t-1})\\
    &\qquad p(\pmb \up_{t-1},\pmb x_{t-1}|\pmb z_{t-1})\,\df\pmb \up_{t-1}d\pmb
    x_{t-1}.
  \end{split}
\end{equation}
Using the Bayes theorem, we can also write,
\begin{equation}
  p(\up_t,x_t|\pmb z_{t-1})= \frac{p(\pmb z_{t-1}|\up_t, x_t)
    \,p(\up_t, x_t)}{\int_{\pmb\Omega_{\up,x}} p(\pmb z_{t-1} |
    \up_t, x_t) \, p(\up_t, x_t)\,\df \up_t\df x_t}.
\end{equation}
Since the noises and states are independent, we can further write,
\begin{equation}
  p(\pmb \up_t, \pmb x_t) = p(\pmb \up_t) \,p(\pmb x_t)= p(\pmb x_t)
  \prod_{l=1}^t p(\up_l).
\end{equation}
Note that the presence of noises $\bup_t$ makes the direct calculation of the
posterior computationally intractable.

In order to enable Bayesian variational inference, we need to define or find
the variational distributions $q$ which can approximate well the exact
distributions $p$. In particular, consider the variational likelihood
$q(\pmb x_t | \pmb z_t)$ and the variational posterior
$q(\pmb \up_t, \pmb x_t|\pmb z_t)$. We have that,
\begin{equation}
  q(\pmb \up_t, \pmb x_t|\pmb z_t) = q(\pmb x_t|\pmb z)
  \,q(\pmb \up_t|\pmb z_t).
\end{equation}
The variational distributions should be selected, so they maximize the
approximation fitness defined as \cite{attias2000variational},
\begin{equation}
  \mathcal{C}_f [q]= \int_{\pmb\Omega_{\up}^t} \df\pmb \up_t \ q(\pmb
  x_t|\pmb z_t)\,q(\pmb \up_t |\pmb z_t)
  \ln\frac{p(\pmb x_t | \pmb z_t)\,p(\pmb \up_t | \pmb z_t)}{q(\pmb x_t|\pmb
    z_t)\, q(\pmb \up_t |\pmb z_t)}.
\end{equation}
The fitness function can be rewritten using the Kullback-Leibler (KL) distance
as \cite{hu2013kullback},
\begin{equation}\label{eq:30}
  \mathcal{C}_f [q]= \E{ \ln\frac{p(\pmb x_t | \pmb z_t)\,p(\pmb \up_t | \pmb
      z_t)}{q(\pmb x_t|\pmb z_t)\, q(\pmb \up_t |\pmb z_t)}}
  - \mathrm{KL}\left[q(\pmb \up_t)\|p(\pmb \up_t)\right]
\end{equation}
where the expectation is taken with respect to the distribution
$q(\pmb \up_t, \pmb x_t|\pmb z_t)$. The expectation represents the mean
likelihood, and the second term in \eref{eq:30} is the KL distance between the
variational posterior and the true posterior of $\pmb \up_t$. Furthermore,
using Jensen's inequality, the fitness function can be upper-bounded for any
variational distributions $q$ by the marginal likelihood, i.e.,
\cite{attias2000variational, neal1998view}
\begin{equation}
  \mathcal{C}_f [q]\leq \ln p(\pmb z_t|\pmb x_t)p(\pmb z_t|\pmb \up_t).
\end{equation}
Therefore, the best variational posterior density must satisfy,
\begin{equation}
  \frac{\df}{\df q(\pmb x_t)} \mathcal{C}_f [q] = 0
\end{equation}
subject to $q(\pmb x_t)\geq 0$. Since the states $\pmb x_t$ are multivariate
Gaussian distributed, we get,
\begin{equation}
  q(\pmb x_t|\pmb z_t)\propto \exp\left(\EE{\pmb\up_t}
    {p(\pmb z_t|\pmb x_t)p(\pmb z_t|\pmb \up_t)}\right)
\end{equation}
where the expectation is calculated with respect to the distribution
$q(\pmb \up_t)$. The variational log-likelihood can be then computed as,
\begin{equation}
  \ln q(\pmb x_t|\pmb z_t) \propto \int_{\pmb\Omega_\up^t}\!\! \df \pmb \up_t\,
  G(\pmb \up_t|\alpha,\beta) \ln p(\pmb z_t|\pmb x_t)p(\pmb z_t|\pmb \up_t).
\end{equation}
Recall that $\pmb x_t\sim N(\pmb x_t;\pmb C_{\pmb x_t})$ (i.e., the zero-mean
multivariate Gaussian distribution with the covariance matrix
$\pmb C_{\pmb x_t}$) and $\pmb \up_t \sim G(\pmb \up_t; \alpha,\beta)$ (i.e.,
the product of $t$ gamma distributions). In addition, to obtain the variational
distribution $q(\pmb x_t)$, we replace the likelihood distribution
$p(\pmb z_t | \pmb \up_t)$ with the independent Gaussian samples
$N(z_l|\up_l)$, and get,
\begin{equation}
  \begin{split}
    \ln q(\pmb x_t|\pmb z_t) =
    &\int_{\pmb \Omega_\up^t} \df\pmb \up_t \ G(\up_t;\alpha,\beta)\\
    &\biggl\{ \ln N(\pmb x_t;\pmb C_{\pmb x_t})
    + \ln G(\pmb \up_t; \alpha,\beta) \\
    &+ \sum_{l=1}^t \ln N(z_l | \up_l) \biggr\}.
  \end{split}
\end{equation}
This expression can be further manipulated as,
\begin{equation}
  \begin{split}
    \ln q(\pmb x_t|\pmb z_t) \propto
    & \ln N(\pmb x_t;\pmb C_{\pmb x_t}) \\
    &+\sum_{l=1}^t \int_{\pmb \Omega_\up} \!\! \df \up_l\, \biggl\{ \ln
    G(\up_l; \alpha,\beta)+ \ln N(z_l | \up_l) \biggr\}.
  \end{split}
\end{equation}
Using the substitution,
\begin{equation}
  ln(z_l|x_l)\propto K^{te}+ (z_l-x_l)^2\E{\up_l}
\end{equation}
for some constant $K$, we can simplify the variational log-distribution as,
\begin{equation}
  \begin{split}
    \ln q(\pmb x_t|\pmb z_t) \propto 
    &\ln N(\pmb x_t; \pmb C_{\pmb x_t}) \\
    &+\sum_{l=1}^t (z_l-x_l)^2\E{\up_l} + \ln K^{-te}.
  \end{split}
\end{equation}
Consequently, the variational distribution becomes,
\begin{equation}
  q(\pmb x_t|\pmb z_t) \propto
  K^{-te}\,N(\pmb x_t; \pmb C_{\pmb x_t}) 
  \exp\!\left\{\sum_{l=1}^t (z_l-x_l)^2\E{\up_l}\right\}.
\end{equation}
This variational distribution can be used in the SMC-SISR method to generate
samples representing the posterior distribution
$p(x_t|\pmb x_{t-1},\pmb z_{1:t})$.

\section{Variational Bayesian estimation}

We adopt a non-parametric inference strategy involving random sampling. In
particular, assuming the distributions derived in the previous section, we
modify the SMC-SISR estimator to fit our scenario of estimating a hidden ARMA
random process from non-linear and noisy observations. The SMC-SISR estimator
represents the posterior density by a set of evolving random samples and the
associated weights. The efficiency of this estimator is strongly influenced by
the choice of the sampling distribution \cite{mihaylova2005particle,
  mihaylova2014overview}. In many cases, using the prior distribution as the
sampling distribution can be a sensible choice \cite{casarin2004bayesian}.

Consider the set of $N$ particles and their weights, i.e.,
$\left\{x_t^{(i)},w_t^{(i)}\right\}_{i=1}^{N}$, evolving over a discrete time
$t$ to represent the posterior distribution $p(\pmb x_t|\pmb z_t)$. The weights
are periodically normalized, so that, $\sum_{i=1}^N w_t^{(i)} = 1$ at all time
instances. The posterior density at time $t$ is approximated as,
\begin{equation}
  p(\pmb x_t|\pmb z_t) \approx \sum_{i=1}^N w_t^{(i)} \delta(\pmb x_t - \pmb
  x_t^{(i)}).
\end{equation}
Thus, the distances $\delta$ between the $i$-th particle $\pmb x_t^{(i)}$ and
the true latent states $\pmb x_t$ are weighted by the coefficients $w_t^{(i)}$.
Here, we assume that particles are generated from the proposal density used in
the importance sampling of the SISR estimator. The proposal distribution can be
represented recursively as,
\begin{equation}
  \pi(x_t|\pmb x_{t-1},\pmb z_t) = \pi(x_1|z_1) \prod_{l=2}^{t} \pi(x_l|\pmb
  x_{l-1},\pmb z_l).
\end{equation}
Consequently, the weight for the latent state $x_t$ required in designing the
importance sampling is calculated as the ratio of the posterior and the
proposal distributions, i.e.,
\begin{equation}
  w(x_t) = \frac{p(x_t|\pmb x_{t-1},\pmb z_t)}{\pi(x_t|\pmb x_{t-1},\pmb z_t)}.
\end{equation}
It is, however, more convenient to calculate the weights recursively as,
\begin{equation}
  w(x_t) = w(x_{t-1})\,
  \frac{p(z_t|x_t)p(x_t|\pmb x_{t-1})}{\pi(x_t|\pmb x_{t-1},\pmb z_t)}.
\end{equation} 
The weights are then normalized as,
\begin{equation}
  \tilde{w}(x_t^{(i)}) =
  \frac{w(x_t^{(i)})\frac{p(z_t|x_t^{(i)})p(x_t^{(i)}|\pmb x_{t-1}^{(i)})}
    {\pi(x_t^{(i)}|\pmb x_{t-1}^{(i)},\pmb z_t)}}{\sum_{i=1}^N   
    w(x_t^{(i)})\frac{p(z_t|x_t^{(i)})p(x_t^{(i)}|\pmb x_{t-1}^{(i)})}
    {\pi(x_t^{(i)}|\pmb x_{t-1}^{(i)},\pmb z_t)}}.
\end{equation}
Since in our case, the proposal distribution is equal to the prior
distribution, the normalized weights can be approximated as,
\begin{equation}
  \tilde{w}(x_t^{(i)}) = w(x_{t-1}^{(i)})\,p(z_t|x_t^{(i)}).
\end{equation}

Furthermore, to resolve the degeneracy of samples which occurs with all
sequential sampling methods, the resampling step at every iteration discards
the particles with small weights, and replace them with new particles having
larger weights \cite{doucet2009tutorial}. More specifically, at each time step,
the $i$-th particle is replaced with the probability $1-\tilde{w}(x_t^{(i)})$
by a new particle. If the particle is replaced, the new particle is assigned
the weight equal to the arithmetic mean of the other particles
\cite{martino2016, martino2018}. Selecting the proper particle weight is
important to yield an unbiased SMC estimation of the marginal likelihood, and
to maintain the particle diversity.

\section{Numerical examples}

Numerical examples are used to validate and evaluate the estimation accuracy of
the devised SMC-SISR estimator utilizing the derived distributions and the
sample weights. The examples assume stochastic log-volatility state-space model
from \cite{casarin2004bayesian}. The volatility model has many applications in
finance and econometrics where it is used to assess the risks
\cite{men2012bayesian}. Mathematically, the stochastic volatility model is an
example of the non-linear ARMA model defined in \eref{eq:10}. The volatility
model generates the zero-mean stationary fractional Gaussian states $x_t$ with
the observations described as,
\begin{equation}\label{eq:45}
  z_t=\up_t \ \eee^{x_t/2}.
\end{equation}

Provided that $\hat{x}_t$ denotes the estimate of the true state $x_t$, the
estimation accuracy can be evaluated as the root mean-square error (RMSE)
defined as,
\begin{equation*}
  \mathrm{RMSE}_t = \sqrt{\frac{1}{t}\sum_{i=1}^{t} (x_i - \hat{x}_i)^2 }.
\end{equation*}

The ARMA models and their parameters used in our numerical experiments are
summarized in Table 1. The variance of the innovation process $u_t$ is set to
unity, i.e., $\sigma_u^2=1$. The gamma distribution of the IID observation
noises has the parameters $\alpha=1/2$ and $\beta=1$. The number of particles
tracked by the SMC-SISR estimator is $N=1000$. We did not observe any
noticeable improvement in the estimation accuracy for larger values of $N$.
However, it is possible to fine tune the effective number of particles to be
less than this value. More importantly, our numerical experiments showed that
the estimation accuracy of the low-order ARMA models is comparable with the
estimation accuracy of the higher-order ARMA models.

\begin{table}[width=.9\linewidth,cols=7,pos=t]
  \caption{The ARMA models used in simulations presented in Figures
    2--7.}\label{tbl1}
  \begin{tabular*}{\tblwidth}{@{} CCLLLLL@{} }
    \toprule
    Figure & Model & $\phi_1$ & $\phi_2$ & $\varphi_1$ & $\varphi_2$ & $H$\\
    \midrule
    2 & ARMA$(1,1)$ & $0.85$ & - & $0.8$ & - & $0.7$ \\	
    3 & ARMA$(2,1)$ & $0.49$ & $0.49$ & $0.8$ & - & $0.8$ \\
    4 & AR$(1)$ & $0.6$ & - & - & - & $0.7$ \\
    5 & MA$(1)$ & - & - & $0.5$ & - & $0.7$ \\
    6 & AR$(2)$ & $0.49$ & $0.45$ & - & - & $0.8$ \\
    7 & MA$(2)$ & - & - & $0.49$ & $0.47$ & $0.8$ \\
    \bottomrule
  \end{tabular*}
\end{table}

The simulation results are presented in Figures 2 -- 7. Each figure consists of
two parts. The upper plot compares a sample realization of the true latent
state sequence $x_t$ with the estimated sequence $\hat{x}_t$. The lower plot
then shows the corresponding RMSE of the estimated sequence.

\begin{figure}
  \centering
  \begin{minipage}{.5\textwidth}
    \centering
    \hspace*{-0.5cm}\includegraphics[scale=0.65]{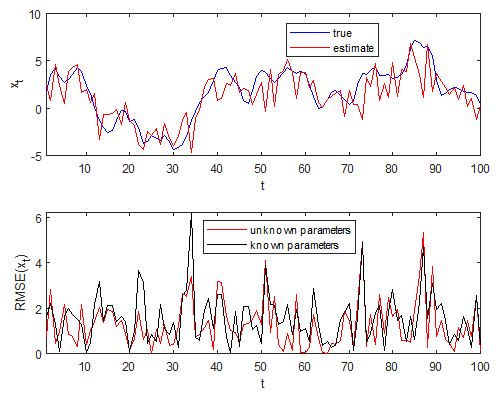}
  \end{minipage}	
  \caption{The sample latent sequence $x_t$ and its estimate $\hat{x}_t$ for
    the ARMA$(1,1)$ model and the corresponding RMSE.}
  \label{fig2}
\end{figure}
\begin{figure}
  \centering
  \begin{minipage}{.5\textwidth}
    \centering
    \hspace*{-0.5cm}\includegraphics[scale=0.65]{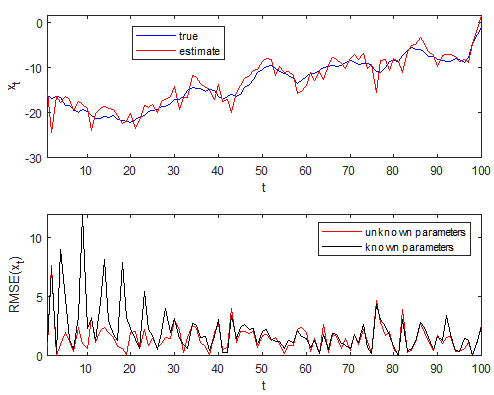}
  \end{minipage}
  \caption{The sample latent sequence $x_t$ and its estimate $\hat{x}_t$ for
    the ARMA$(2,1)$ model and the corresponding RMSE.}
  \label{fig3}
\end{figure}

\begin{figure}
  \centering
  \begin{minipage}{.5\textwidth}
    \centering
    \hspace*{-0.5cm}\includegraphics[scale=0.65]{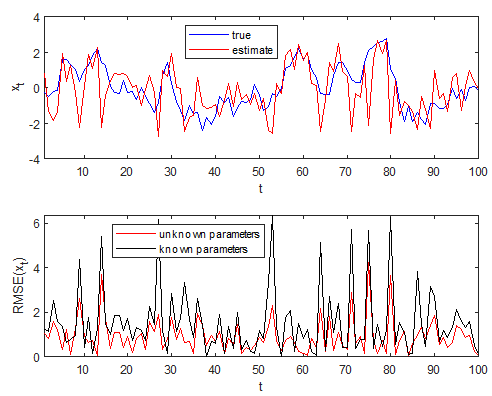}
  \end{minipage}
  \caption{The sample latent sequence $x_t$ and its estimate $\hat{x}_t$ for
    the AR$(1)$ model and the corresponding RMSE.}
  \label{fig4}
\end{figure}

\begin{figure}
  \centering
  \begin{minipage}{.5\textwidth}
    \centering
    \hspace*{-0.5cm}\includegraphics[scale=0.65]{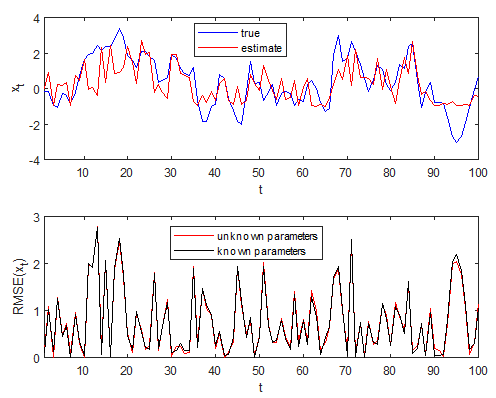}
  \end{minipage}
  \caption{The sample latent sequence $x_t$ and its estimate $\hat{x}_t$ for
    the MA$(1)$ model and the corresponding RMSE.}
  \label{fig5}
\end{figure}

\begin{figure}
  \centering
  \begin{minipage}{.5\textwidth}
    \centering
    \hspace*{-0.5cm}\includegraphics[scale=0.65]{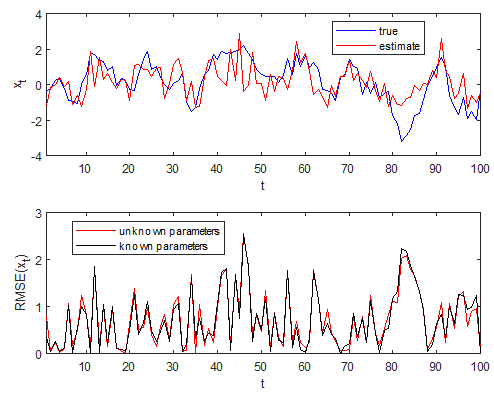}
  \end{minipage}
  \caption{The sample latent sequence $x_t$ and its estimate $\hat{x}_t$ for
    the AR$(2)$ model and the corresponding RMSE.}
  \label{fig6}
\end{figure}

\begin{figure}
  \centering
  \begin{minipage}{.5\textwidth}
    \centering
    \hspace*{-0.5cm}\includegraphics[scale=0.65]{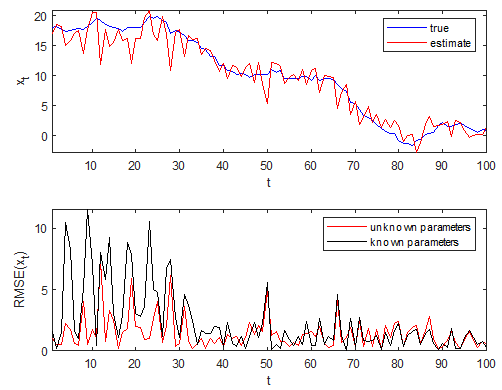}
  \end{minipage}
  \caption{The sample latent sequence $x_t$ and its estimate $\hat{x}_t$ for
    the MA$(2)$ model and the corresponding RMSE.}
  \label{fig7}
\end{figure}

\subsection{Gene expression time series example}

We now investigate more complex as well as more practical problem of inferring
missing values in the gene expression time series. The inference is performed
by the SMC-SISR estimator developed in the previous sections. The missing data
in biological experiments may be caused by unexpectedly large measurement
errors (outliers), or due to other intermittent problems in the experimental
measurements. The gene expression data are vital for reconstructing gene
expression networks \cite{yao2015prior}. More importantly, simple deterministic
interpolation to replace the missing data values did not provide satisfactory
results \cite{bar2003continuous}.

The gene expression data can be modeled by a vector autoregressive (VAR) model
\cite{fujita2007modeling, tam2013synthetic}. We can directly employ the
SMC-SISR estimator instead of first estimating the parameters of this model, In
particular, the sequence $\pmb x_t$ of the gene expression time series can be
modeled recursively by the $k$-order VAR model, i.e., let,
\begin{equation}\label{eq:50}
  \pmb x_t=\sum_{i=1}^{k} \pmb W_i\,  \pmb x_{t-i}+\pmb u_t
\end{equation}
where $n$ is the number of genes (i.e., the sample size of the time series data
observed at each time instant), the $(n\times n)$ matrices $\pmb W_i$ represent
the model parameters, and $\pmb u_t$ is a $(n\times 1)$ vector of Gaussian
innovations which are zero mean and uncorrelated. The observations are modeled
as in \eref{eq:45}.

The actual gene expression data for the Hela cell were obtained from the
supplementary material provided in \cite{whitfield2002identification}. The data
provide measurements at $16$ time instances separated by one hour over $3$ cell
cycles, so the total number of data points is $48$. The data for the genes 7p21
and p53 were selected as two representative examples. Rather than fitting the
model parameters to the model described by \eref{eq:45} and \eref{eq:50}, we
can use the SMC-SISR estimator with $1000$ particles to track the most probable
counts of the gene produced in the cell. The actual and predicted counts of the
two genes are shown in Figure 8 and 9, respectively. We can observe that the
random sampling model follows the expression data reasonably well, and the
estimation error is likely sufficient for many biological applications.

\begin{figure}
  \centering
  \begin{minipage}{.5\textwidth}
    \centering
    \hspace*{-0.5cm}\includegraphics[scale=0.68]{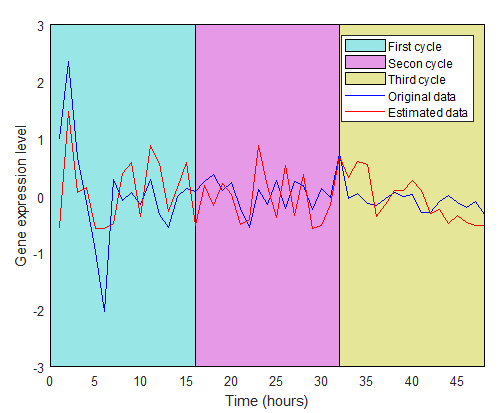}
  \end{minipage}	
  \caption{Tracking the expression data for the 7p21 gene in the Hela cell
    using the SMC-SISR estimator.}
  \label{fig11}
\end{figure}

\begin{figure}
  \centering
  \begin{minipage}{.5\textwidth}
    \centering
    \hspace*{-0.5cm}\includegraphics[scale=0.68]{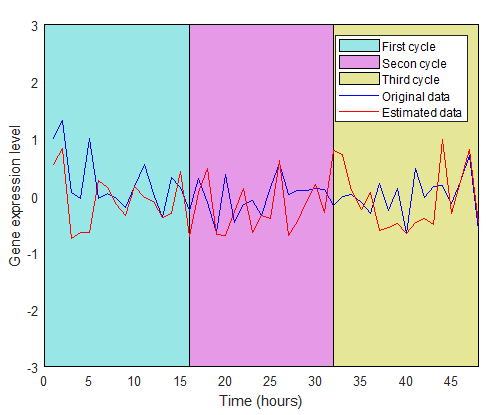}
  \end{minipage}
  \caption{Tracking the expression data for the p53 gene in the Hela cell using
    the SMC-SISR estimator.}
  \label{fig12}
\end{figure}

\section{Discussion}

Considering numerical experiments and other results presented in this paper, we
can conclude that the latent time series can be estimated with good accuracy
from non-linear and noisy observations even if the parameters of the underlying
model are not known. The unknown parameters include the ARMA model coefficients
as well as the parameters describing the input innovation process such as the
variance, and the autocorrelation. Such assumption may be important especially
in situations when all or some of the model parameters cannot be easily
estimated. For example, the parameters may be time varying and even
non-stationary \cite{bibi2006evolutionary, pintelon1999time}. In such case, the
SMC estimator may not be able to track the parameter variations. However, if
the model parameters vary sufficiently fast, the SMC estimator can instead
follow possibly time-varying mean values of the model parameters representing
slowly changing random processes.

Provided that the latent states are non-negative integers or positive real
values, the measurement noise is generally dependent on the latent state values
in order to satisfy this constraint. The correlations between the states and
the measurement noises would substantially complicate the state inference. In
this paper, we assumed that the measurement noises are gamma distributed and
IID, so it does not violate the non-negativity constraint of the latent state
values. Furthermore, the computational complexity of calculating the posterior
of latent states dictates the use of variational Bayesian inference or other
methods for approximating the posterior distribution \cite{beal2003variational,
  bracegirdle2013inference}.

The performance of the SMC-SISR estimator was shown to be good for several
low-order ARMA models. There are many practical applications where the
inference of latent states from observations is important. For instance, such
inferences are used to reconstruct the state space models of dynamic systems
\cite{casdagli1991state, yang2012spatiotemporal}, or as illustrated in this
paper, we can infer missing values in the time series data. The gene expression
data are an example of the multi-dimensional time series where the expressions
of multiple genes are measured in parallel at discrete time instances. In
general, gene expression data are useful to reconstruct gene reaction networks,
and to elucidate the properties and understanding of genetic circuits
\cite{atitey2018determining, atitey2018inferring, atitey2019elucidating}.

\section{Conclusion}

The SMC sampling estimators are usually used to perform variational Bayesian
inference in order to reduce the computational complexity by approximating the
posterior distribution. More importantly, these estimators such as the SMC-SISR
estimator adopted in this paper can be effective in estimating the hidden
random processes from non-linear and noisy observations, even if the parameters
of the underlying state space model are not known. The numerical results
indicate that the SMC-SISR estimator achieves good estimation accuracy,
especially for the low-order ARMA models, and this estimator is also unbiased.

There are many practical situations where the inference problem considered in
this paper is encountered. One such problem was briefly investigated to
demonstrate the performance of the SMC-SISR estimator for the time series data
with multiple observations at each time instant to infer the missing values.
Future work will consider different non-linearity observation functions and
noise distributions, and the inference problem for the data models with
time-varying random parameters. In this latter case, the SMC sampling
estimators may track changes in the hidden state statistics, or these
statistics can be averaged out from the likelihood function or from the
posterior distribution.

\bibliographystyle{plain}
\bibliography{References}

\end{document}